\documentclass{article}

\usepackage{microtype}
\usepackage{graphicx}
\usepackage{subcaption}
\usepackage{booktabs} % for professional tables
\usepackage{tabularx}
\usepackage{array}
\usepackage{amsmath}
\usepackage{amssymb}
\usepackage{mathtools}
\usepackage{tikz}
\usepackage{hyperref}

\usepackage[preprint]{icml2027}

\usepackage{amsmath}
\usepackage{amssymb}
\usepackage{mathtools}
\usepackage{amsthm}

\usepackage[capitalize,noabbrev]{cleveref}

\theoremstyle{plain}

\theoremstyle{definition}

\theoremstyle{remark}

\usepackage[textsize=tiny]{todonotes}

\icmltitlerunning{Falsifiable Compositional Reasoning in Time Series Forecasting}

\begin{document}

\twocolumn[
  \icmltitle{Falsifiable Compositional Reasoning in Time Series Forecasting}

  % the [accepted] option to the icml2027 package.


  \icmlsetsymbol{equal}{*}

  \begin{icmlauthorlist}
    \icmlauthor{Hao Xue}{sch}
  \end{icmlauthorlist}

  \icmlaffiliation{sch}{Hong Kong University of Science and Technology (Guangzhou), Guangzhou, China}

  \icmlcorrespondingauthor{Hao Xue}{haoxue@hkust-gz.edu.cn}


  \vskip 0.3in
]



\printAffiliationsAndNotice{}  % no special notice (required even if empty)

\begin{abstract}
Time series foundation models as well as the language-guided analytical agents achieve strong accuracy on standard chronological benchmarks. However, models that rank highest on historical validation splits often degrade across non-stationary regime transitions. We hypothesise that this gap arises from using historical backtesting as an open-ended search and model-selection objective instead of using it to test pre-specified compositional assumptions. Real-world time series are generated by recurring latent primitives that co-move within finite historical windows. Under strong historical co-movement, chronological holdout error cannot separate predictors that learn invariant couplings among primitives from predictors that fit transient historical correlations. Searching over large pools of architectures, hyperparameters, or chain-of-thought rules further biases selection toward historical shortcuts. Through theoretical analysis and diagnostic experiments on a controlled multi-primitive benchmark and German day-ahead electricity prices, we show that expanding validation search fails to improve out-of-regime accuracy, and that decomposing primitives without an ex-ante falsification test still breaks down when residual correlations shift. Hence, we propose formulating time series forecasting and reasoning around falsifiable compositional hypotheses that combine neural primitive encoders, explicit coupling constraints, and training-calibrated falsification tests.

\end{abstract}
\section{Introduction}
\label{sec:intro}
% =============================================================================

Recent progress in time series forecasting has followed three main directions. Task-specific neural architectures, including patch and channel Transformers \citep{nie2023patchtst,liu2024itransformer} as well as basis and trend-seasonal decompositions \citep{oreshkin2020nbeats,wu2021autoformer,zeng2023dlinear}, have improved accuracy and computational efficiency on long-horizon benchmarks. Pretrained time series foundation models (TSFMs) fitted on multi-domain corpora have extended these sequence backbones to zero-shot and few-shot forecasting \citep{ansari2024chronos,das2024timesfm,woo2024moirai,goswami2024moment,shi2025timemoe}. At the same time, language-guided and multimodal systems have begun combining numerical inputs with domain text and explicit reasoning steps \citep{jin2024timellm,jin2024position,williams2025context}. On standard chronological train-validation-test splits, these approaches fit historical patterns accurately.

When models move from static benchmark splits to non-stationary deployments, however, high historical validation accuracy frequently fails to translate into out-of-sample stability. Recent empirical studies have documented several facets of this discrepancy. Forecasters and TSFMs that extrapolate individual basis signals accurately degrade on unseen combinations of those same signals \citep{potosnak2025compositional}, and multi-step autoregressive rollouts often repeat input context motifs without recovering underlying state-space dynamics \citep{durstewitz2026dynamical}. Large-scale benchmark re-evaluations also show that performance differences among neural architectures largely disappear once hyperparameter tuning budgets are matched \citep{brigato2025position,moretti2026benchmarking}, while supervised fine-tuning of multimodal time series reasoners can reduce out-of-domain compositional transfer \citep{merrill2024language}. 
These findings suggest that increasing model capacity or improving historical fit does not necessarily resolve the problem of generalisation under structural shifts.

\begin{figure*}[t]
  \centering
  \includegraphics[width=0.95\textwidth]{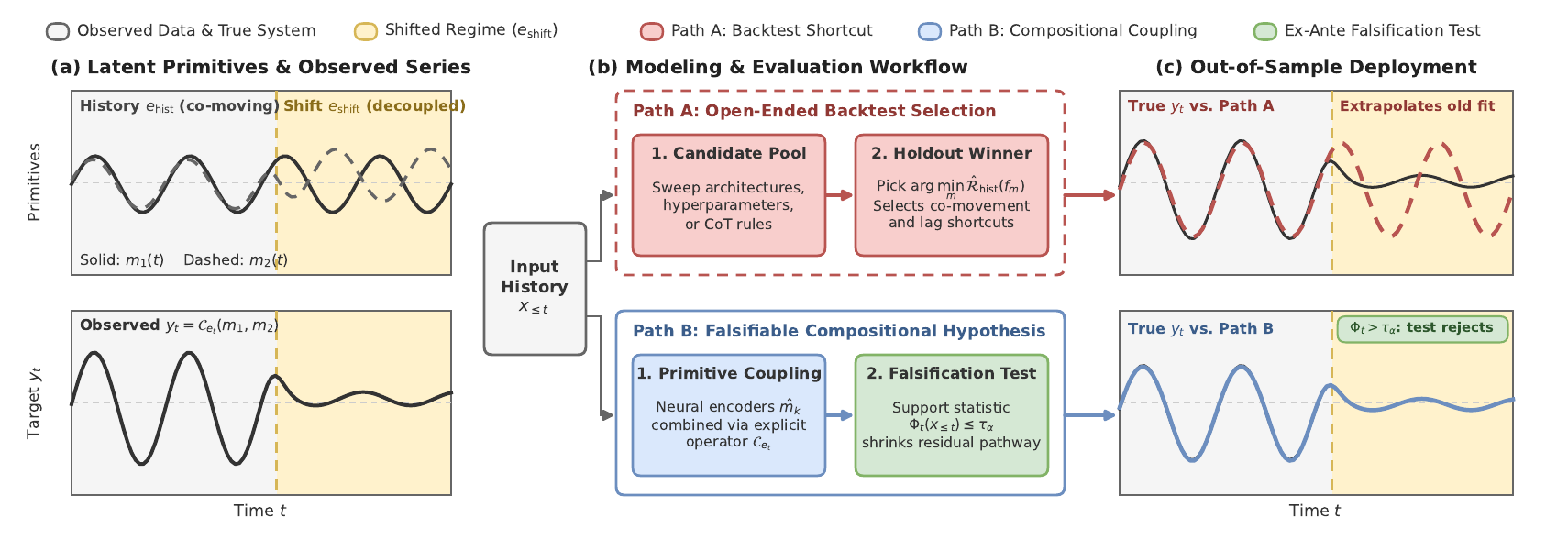}
  \caption{\textbf{Open-ended backtest selection versus falsifiable compositional reasoning.} \textbf{(a)} Observed series $y_t$ arise from latent primitives $m_1(t), m_2(t)$ that co-move in the historical regime $e_{\mathrm{hist}}$ and decouple in $e_{\mathrm{shift}}$. \textbf{(b)} Path A searches candidate pools to minimise historical holdout error, rewarding historical co-movement shortcuts, whereas Path B combines neural primitive encoders $\hat{m}_k$ under an explicit operator $\mathcal{C}_{e_t}$ with a training-calibrated falsification test $\Phi_t(x_{\leq t}) \leq \tau_\alpha$. \textbf{(c)} Under $e_{\mathrm{shift}}$, Path A extrapolates the outdated correlation, while Path B rejects the shifted residual fit and preserves the structural coupling.}
  \label{fig:overview}
\end{figure*}

Hence, in this work, we hypothesise that \textit{historical backtesting cannot identify compositional mechanisms when treated as an open-ended search or model-selection objective. Reliable generalisation across regime shifts requires formulating time series models as \textbf{falsifiable compositional hypotheses} that specify grounded temporal primitives, explicit coupling constraints, and pre-committed ex-ante falsification tests prior to holdout evaluation.}
We argue that these failure modes stem from a shared evaluation practice: using chronological backtesting as an open-ended discovery and selection objective (Figure~\ref{fig:overview}). Time series in physical, economic, and clinical systems are generated by recurring temporal primitives, such as daily load cycles, weather-driven generation, or inventory mean-reversion, coupled through regime-dependent constraints. Over any finite historical period, distinct primitives often co-move in stable proportions. As long as that collinearity persists, an unconstrained model that fits the joint historical trajectory achieves nearly the same chronological validation loss as a model that isolates individual primitives and their coupling rule. Chronological error alone therefore leaves compositional structure unidentifiable. When practitioners or automated agents search over many candidate architectures, lookback lengths, or reasoning rules to minimise validation loss, multiple-testing selection bias systematically favours candidates that fit transient historical correlations \citep{bailey2014deflated,bailey2014pbo}.

Addressing this problem requires separating representation learning from hypothesis testing. Before evaluating on a historical holdout split, a compositional forecast specifies the underlying temporal primitives, the directional constraints governing their coupling under observable regime boundaries, and a quantitative ex-ante test that suppresses local residual adjustments when input relationships leave training support. Deep sequence encoders and pretrained TSFMs serve as modular estimators that extract noisy primitives from raw observations, analogous to factorised generative modules in vision and decision-making \citep{lake2017building,du2024compositional}. Pairing these neural encoders with constrained composition layers and ex-ante falsification tests preserves predictive accuracy in stationary periods while maintaining invariance when familiar primitives recombine in new proportions.

\section{Why Historical Backtesting Cannot Identify Compositional Structure}
\label{sec:theory}
% =========================
To analyse why validation improvements on historical splits often vanish after regime shifts, we formalise how latent temporal primitives combine to generate observed sequences and characterise the set of predictors minimised by chronological error.

\subsection{Time Series as Regime-Conditioned Compositions of Primitives}
\label{sec:theory_dgp}

Let $x_{\leq t} = (x_{t-L+1}, \dots, x_t) \in \mathbb{R}^{L \times d}$ denote the observed history of a target variable and exogenous covariates up to step $t$, and let $y_{t+1:t+H} \in \mathbb{R}^H$ denote the forecast target over horizon $H$. In many domains, $y_{t+1:t+H}$ is generated by $K$ recurring latent primitives $m_1(x_{\leq t}), \dots, m_K(x_{\leq t})$ combined through a regime-dependent composition operator $\mathcal{C}_{e_t}$:
\begin{equation}
  y_{t+1:t+H} = \mathcal{C}_{e_t}\!\bigl(m_1(x_{\leq t}), \dots, m_K(x_{\leq t})\bigr) + \varepsilon_{t+1:t+H},
  \label{eq:cdgp}
\end{equation}
where $e_t \in \mathcal{E}$ denotes the active regime and $\varepsilon_{t+1:t+H}$ is zero-mean noise with variance $\sigma_\varepsilon^2$. Each primitive $m_k$ captures a reusable temporal component, such as multi-scale seasonal demand, exogenous supply variation, or inventory feedback, while $\mathcal{C}_{e_t}$ specifies how those components add, interact, or clip under capacity and institutional rules \citep{koller2009probabilistic,wiedemer2024compositional,defelice2026interpretability}. During a compositional regime shift from $e_{\mathrm{hist}}$ to $e_{\mathrm{shift}}$, individual primitives $m_k$ remain within their historical marginal support, while their joint correlation structure or the boundary parameters of $\mathcal{C}_{e_t}$ change.

Day-ahead electricity pricing provides a clear example of Equation~\ref{eq:cdgp} \citep{weron2014electricity,lago2021epf}. Hourly market-clearing prices depend on consumer demand cycles ($m_1$), weather-dependent wind and solar supply ($m_2$), and thermal dispatch cost curves ($\mathcal{C}_{e_t}$), which steepen near generation capacity limits. Consumer demand schedules and turbine power curves remain the same physical processes across years. As installed renewable capacity grows, however, the historical correlation between afternoon demand peaks and renewable output changes substantially, shifting the joint covariate distribution while preserving the market-clearing relationship that maps net residual load ($m_1 - m_2$) to price.

\begin{figure*}[t]
  \centering
  \includegraphics[width=0.89\textwidth]{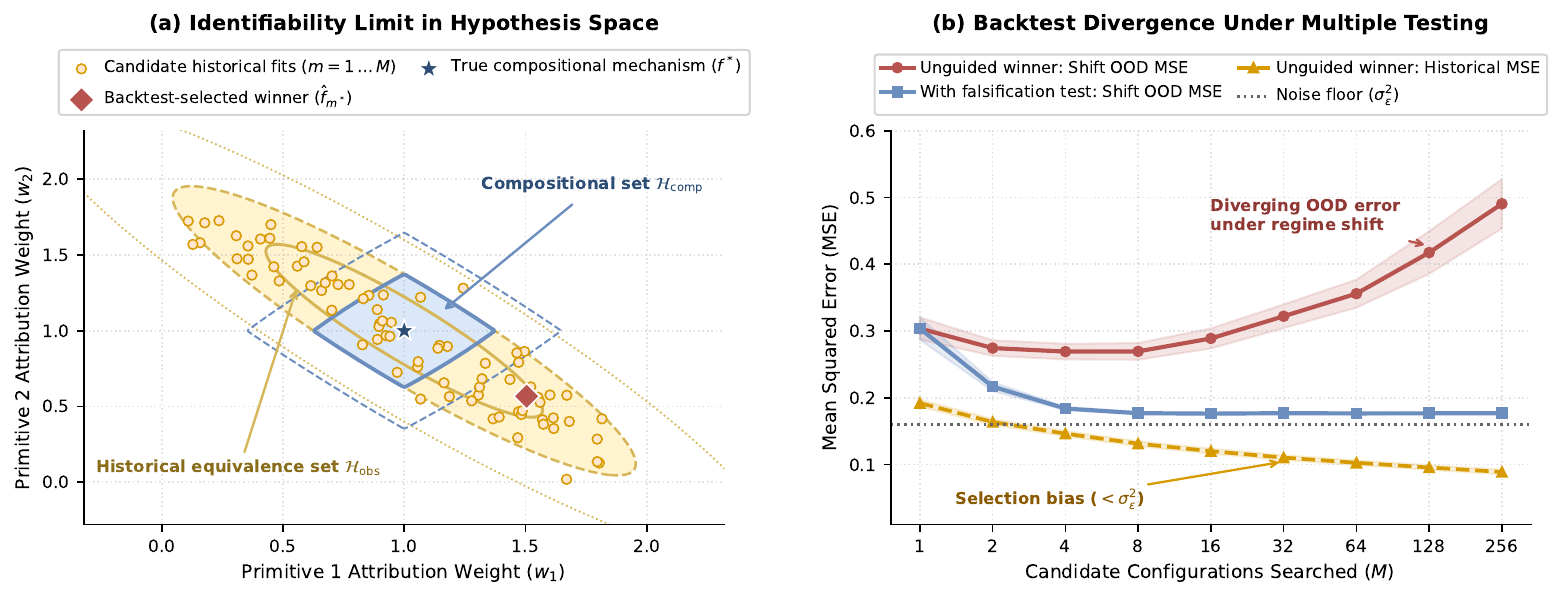}
  \caption{\textbf{Non-identifiability in parameter space and selection bias across candidate pools.} \textbf{(a)} When primitives co-move in the historical regime, chronological risk $\mathcal{R}_{e_{\mathrm{hist}}}(f)$ forms an elongated equivalence region $\mathcal{H}_{\mathrm{obs}}$ containing misattributed predictors alongside the true mechanism $f^*$. Validation noise can push the empirical winner $\hat{f}_{m^*}$ outside the compositionally consistent set $\mathcal{H}_{\mathrm{comp}}$. \textbf{(b)} As search breadth $M$ increases, the unguided winner's historical MSE falls below the noise floor $\sigma_\varepsilon^2$ while its shift MSE rises, whereas filtering candidates with an ex-ante falsification test stabilises out-of-regime error.}
  \label{fig:geometry}
\end{figure*}

\subsection{Observational Indistinguishability Under Historical Co-movement}
\label{sec:theory_identifiability}

Suppose candidate predictors $f \in \mathcal{F}$ are evaluated by expected squared error in the historical regime $e_{\mathrm{hist}}$:
\begin{equation}
  \mathcal{R}_{e_{\mathrm{hist}}}(f) = \mathbb{E}_{e_{\mathrm{hist}}}\!\Bigl[\bigl\|y_{t+1:t+H} - f(x_{\leq t})\bigr\|_2^2\Bigr].
  \label{eq:hist_risk}
\end{equation}
Consider two unit-variance primitives combined linearly as $y = w_1^* m_1 + w_2^* m_2 + \varepsilon$ with historical correlation $\rho_{\mathrm{hist}} = \mathrm{Corr}_{e_{\mathrm{hist}}}(m_1, m_2)$. A linear predictor $f_w(x) = w_1 m_1 + w_2 m_2$ with weight error $\Delta w = w - w^*$ incurs excess historical risk $\Delta w_1^2 + \Delta w_2^2 + 2\rho_{\mathrm{hist}}\Delta w_1 \Delta w_2$. When $m_1$ and $m_2$ are strongly correlated in history ($\rho_{\mathrm{hist}} \to 1$), attribution errors along $\Delta w_1 \approx -\Delta w_2$ produce almost no excess historical loss because $(1 - \rho_{\mathrm{hist}})(\Delta w_1 - \Delta w_2)^2/2 \approx 0$. More generally, for nonlinear operators $\mathcal{C}_{e_{\mathrm{hist}}}$, chronological risk constrains $f$ only on the historical co-occurrence manifold, defining an observational equivalence set
\begin{equation}
  \mathcal{H}_{\mathrm{obs}}(e_{\mathrm{hist}}, \delta) = \bigl\{f \in \mathcal{F} : \mathcal{R}_{e_{\mathrm{hist}}}(f) \leq \sigma_\varepsilon^2 + \delta\bigr\},
  \label{eq:equiv_set}
\end{equation}
whose width orthogonal to the historical manifold expands as primitive collinearity increases (Figure~\ref{fig:geometry}a). Inside $\mathcal{H}_{\mathrm{obs}}$, a model that proxies $m_2$ through lagged targets $y_{t-24}$ or correlated channels achieves nearly the same historical loss as the true composition $f^* = \mathcal{C}_{e_t} \circ (m_1, \dots, m_K)$ \citep{haavelmo1944probability,lucas1976econometric,peters2016causal}.

Finite-sample estimation reinforces this preference for coupled shortcuts. Empirical series typically contain regime-specific serial correlations in $\varepsilon_t$ that do not carry over across structural changes. High-capacity end-to-end networks fit both the structural signal and these local residual correlations jointly, achieving lower chronological validation error in $e_{\mathrm{hist}}$ than a modular predictor restricted to invariant primitives \citep{arjovsky2019invariant}. Minimising Equation~\ref{eq:hist_risk} rewards every predictive correlation present in $e_{\mathrm{hist}}$, whether it arises from a persistent mechanism or a temporary alignment.

Hourly German electricity data from 2012 to 2017 in the EPF-DE benchmark \citep{lago2021epf} ($52{,}416$ hours) illustrates this effect quantitatively. From 2012 to 2014, grid load and combined wind-and-solar generation maintained stable seasonal proportions, yielding a 24-hour price autocorrelation of $r = 0.703$ on validation splits. An unconstrained sequence model achieves low validation error in this period primarily by relying on $y_{t-24}$ and local smoothing. Between 2015 and 2017, wind and solar output increased by $+2.94$ standard deviations (standardised mean shifting from $-0.48$ to $+2.46$), reducing 24-hour price autocorrelation to $r = 0.511$. Over the entire six-year span, however, the 24-hour differenced correlation between price changes $\Delta_{24} y_t$ and net residual-load changes $\Delta_{24}(\mathrm{Load}_t - \mathrm{PVWind}_t)$ remained stable at $0.759$ in training, $0.810$ in validation, $0.786$ in chronological testing, and $0.791$ during the high-renewable period. Once the system enters $e_{\mathrm{shift}}$ ($\rho_{\mathrm{shift}} = -0.75$ in the analytical simulation of Figure~\ref{fig:geometry} across $600$ Monte Carlo trials), the risk contours rotate and separate the compositionally consistent set $\mathcal{H}_{\mathrm{comp}} = \{f : \sup_{e \in \mathcal{E}} \mathcal{R}_e(f) \leq \sigma_\varepsilon^2 + \delta\}$ from shortcuts inside $\mathcal{H}_{\mathrm{obs}}$.

\subsection{Selection Bias When Backtesting Guides Open-Ended Search}
\label{sec:theory_pbo}

Non-identifiability worsens when a single historical validation split is reused to pick a winner among $M$ candidate architectures, hyperparameter settings, or reasoning rules. Let $\hat{\mathcal{R}}_{e_{\mathrm{hist}}}(f_m)$ denote the empirical validation MSE of candidate $m \in \{1, \dots, M\}$ across $T$ windows. Because multi-step forecast errors are serially correlated and grouped by regime, the effective sample size $T_{\mathrm{eff}}$ is much smaller than $T$. Selecting $m^* = \arg\min_{1 \leq m \leq M} \hat{\mathcal{R}}_{e_{\mathrm{hist}}}(f_m)$ across candidates distributed along $\mathcal{H}_{\mathrm{obs}}$ introduces a downward selection bias in apparent validation risk on the order of $\sigma_{\mathrm{val}}\sqrt{(2 \ln M)/T_{\mathrm{eff}}}$ \citep{bailey2014deflated,harvey2016cross,lopezdeprado2018advances}. Candidates with larger attribution errors along the collinear anti-diagonal have higher variance across finite historical slices, making them more likely to win multi-trial backtest comparisons when finite-sample fluctuations align in their favour (Figure~\ref{fig:geometry}b).

The same selection bias applies to language-guided and agentic time series pipelines \citep{dziri2023faith}. Several recent workflows prompt large language models to generate candidate symbolic features, domain explanations, or rule-based adjustments, retaining whichever candidate chain achieves the lowest validation MSE. Searching over natural-language rules obeys the same multiple-testing statistics as searching over hyperparameter grids. Because unconstrained text generation produces a large combinatorial space of plausible post-hoc explanations, ranking generated rules by historical validation MSE searches broadly across $\mathcal{H}_{\mathrm{obs}}$ and selects rules tied to transient historical correlations.

\section{Observational Fit VS. Compositional Generalisation in Practice}
\label{sec:empirical}
% =============================================================================

Equations~\ref{eq:cdgp}--\ref{eq:equiv_set} represent four empirical patterns reported in recent forecasting benchmarks and motivate two diagnostic experiments testing validation selection and ex-ante falsification.

\subsection{Converging Anomalies Across Recent Forecasting Studies}
\label{sec:emp_synthesis}

% Viewed through observational equivalence on $e_{\mathrm{hist}}$, several recent findings follow directly from minimising historical error without compositional constraints. 
Several recent empirical findings can also be interpreted through the observational equivalence framework introduced above.
For example, \citet{potosnak2025compositional} evaluated 16 deep and foundation forecasters on synthetic combinations of trend, periodic, and event functions, finding that models accurate on individual components degrade on unseen pairings unless nearly all pairings appear during training. \citet{brigato2025position} and \citet{moretti2026benchmarking} showed across controlled tuning sweeps that performance gaps among neural architectures largely vanish under equal hyperparameter budgets, consistent with distinct parameterisations reaching similar validation error within $\mathcal{H}_{\mathrm{obs}}$ on quasi-stationary splits.  \citet{durstewitz2026dynamical} also showed that zero-shot TSFMs can obtain low short-horizon MSE by repeating historical motifs while failing to capture underlying dynamical attractors. 

To test the selection and falsification mechanisms from Section~\ref{sec:theory} directly, we run two controlled diagnostic probes on a synthetic multi-primitive benchmark (\texttt{Controlled C-DGP}, $14{,}400$ steps combining diurnal, seasonal-envelope, dispatch, and saturation operators) and the six-year German electricity transition (\texttt{EPF-DE}, $52{,}416$ hourly steps from 2012 to 2017 with horizon $H=24$). The first probe measures how chronological validation rankings relate to out-of-regime error as candidate search breadth $M$ grows. The second probe tests whether decomposing primitives is sufficient on its own or requires a pre-committed ex-ante falsification test.

\begin{figure*}[t]
  \centering
  \includegraphics[width=0.91\textwidth]{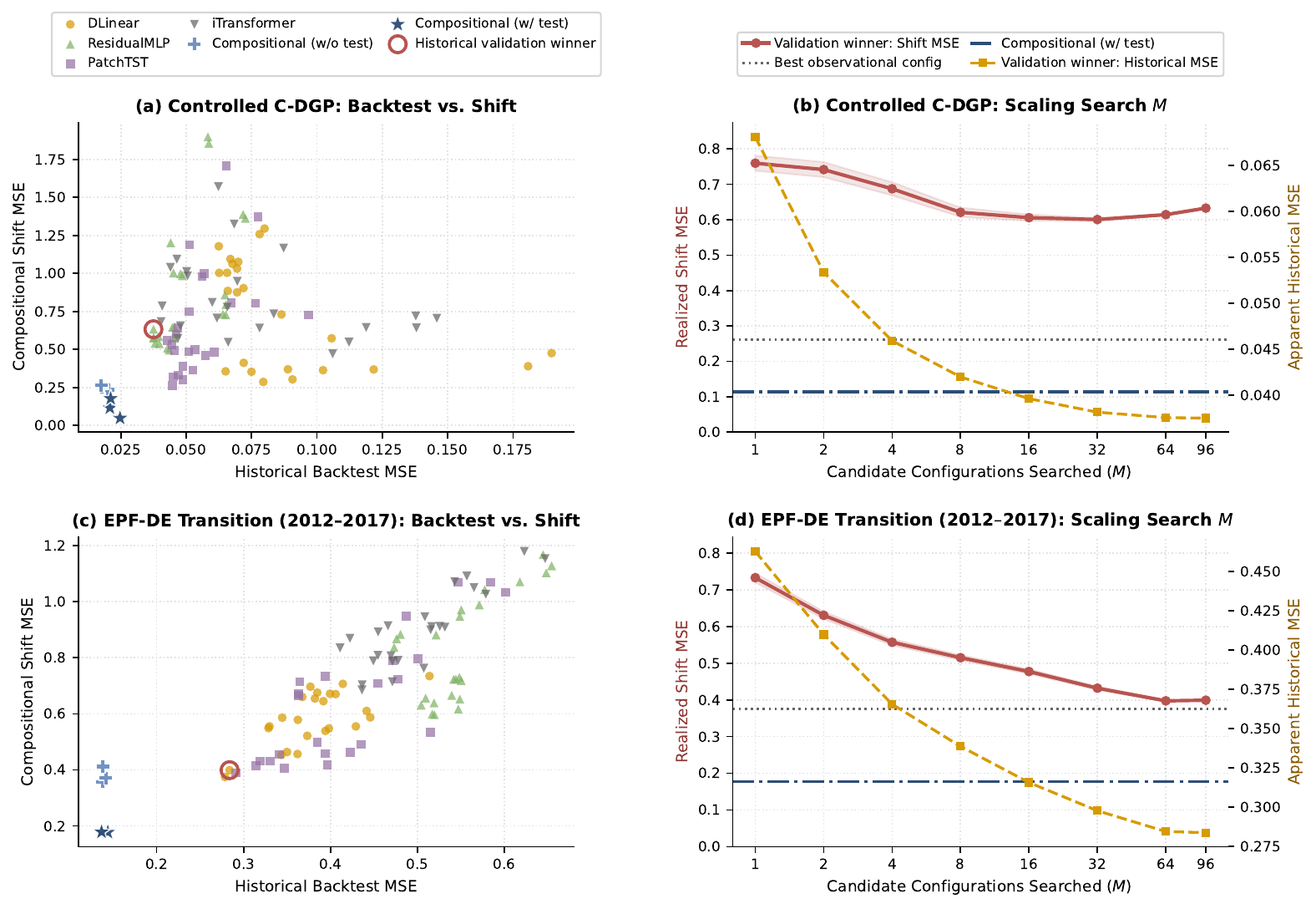}
  \caption{\textbf{Historical backtest ranking versus compositional-shift error.} \textbf{(a, c)} Historical backtest MSE plotted against compositional-shift MSE across four end-to-end backbone families (\texttt{DLinear}, \texttt{ResidualMLP}, \texttt{PatchTST}, \texttt{iTransformer}) and compositional models on \texttt{Controlled C-DGP} and \texttt{EPF-DE}. \textbf{(b, d)} Bootstrap selection curves showing the validation winner's apparent historical MSE and realised shift MSE as search breadth $M$ increases.}
  \label{fig:probe1}
\end{figure*}

\subsection{Scaling Candidate Search Under Compositional Shift}
\label{sec:emp_probe1}

In Diagnostic Probe 1, we evaluate a candidate pool of 96 end-to-end configurations per dataset across four representative backbones: \texttt{DLinear} \citep{zeng2023dlinear}, \texttt{ResidualMLP} \citep{oreshkin2020nbeats}, \texttt{PatchTST} \citep{nie2023patchtst}, and \texttt{iTransformer} \citep{liu2024itransformer}, crossing lookback lengths $L \in \{48, 96, 168\}$, covariate inclusion, reversible instance normalization \citep{kim2022revin}, hidden width, and weight decay across three random seeds. All models are trained on an initial historical period, ranked on a chronological validation split, and evaluated on both a subsequent in-regime test split and a compositional-shift split where latent primitives decouple in phase (\texttt{Controlled C-DGP}) or renewable supply rises by $+2.94$ SD (\texttt{EPF-DE}). For comparison, we evaluate six compositional configurations ($h \in \{24, 32, 48\}$) that combine neural primitive encoders with explicit residual-load coupling, evaluated with and without a training-calibrated falsification test.

Figures~\ref{fig:probe1}a and \ref{fig:probe1}c compare historical backtest error against shift error. On \texttt{Controlled C-DGP} (Figure~\ref{fig:probe1}a), chronological MSE across the 96 observational configurations ranges from $0.037$ to $0.190$, and its Spearman rank correlation with shift MSE is $\rho = +0.077$ ($p = 0.455$). Covariate-coupled \texttt{ResidualMLP} models achieve the lowest validation MSE ($0.0375$) by fitting the historical phase alignment between load and supply, yielding $0.6329$ MSE after decoupling, whereas narrower \texttt{PatchTST} configurations with higher validation error ($0.0446$) obtain $0.2621$ shift MSE. On \texttt{EPF-DE} (Figure~\ref{fig:probe1}c), historical validation separates covariate-aware architectures from univariate baselines, but rank inversions remain across families: \texttt{iTransformer} at $L=168$ achieves lower historical MSE than \texttt{ResidualMLP} while incurring higher shift MSE. Across all 96 observational configurations on \texttt{EPF-DE}, the validation-selected configurations achieve shift MSEs between $0.375$ and $0.399$. Modelling 24-hour residual-load coupling above a 72-hour de-effected baseline with ex-ante falsification reduces mean shift MSE to $0.1135$ on \texttt{Controlled C-DGP} and $0.1774$ on \texttt{EPF-DE}.

Figures~\ref{fig:probe1}b and \ref{fig:probe1}d show bootstrap selection curves ($1{,}000$ draws per $M \in [1, 96]$) when deploying the historical validation winner. On \texttt{Controlled C-DGP} (Figure~\ref{fig:probe1}b), increasing $M$ reduces apparent historical MSE from $0.068$ ($M=1$) to $0.0375$ ($M=96$). Realised shift MSE decreases from $0.760$ at $M=1$ to $0.606$ at $M=16$ as underfitted candidates are removed, and then rises to $0.633$ at $M=96$, where selection misses the observational pool's top-quartile shift models in $100\%$ of trials. On \texttt{EPF-DE} (Figure~\ref{fig:probe1}d), expanding $M$ to $96$ lowers historical MSE to $0.284$ while realised shift MSE remains at $0.399$, more than twice the $0.1774$ error of the falsifiable compositional model.

\begin{figure*}[t]
  \centering
  \includegraphics[width=0.95\textwidth]{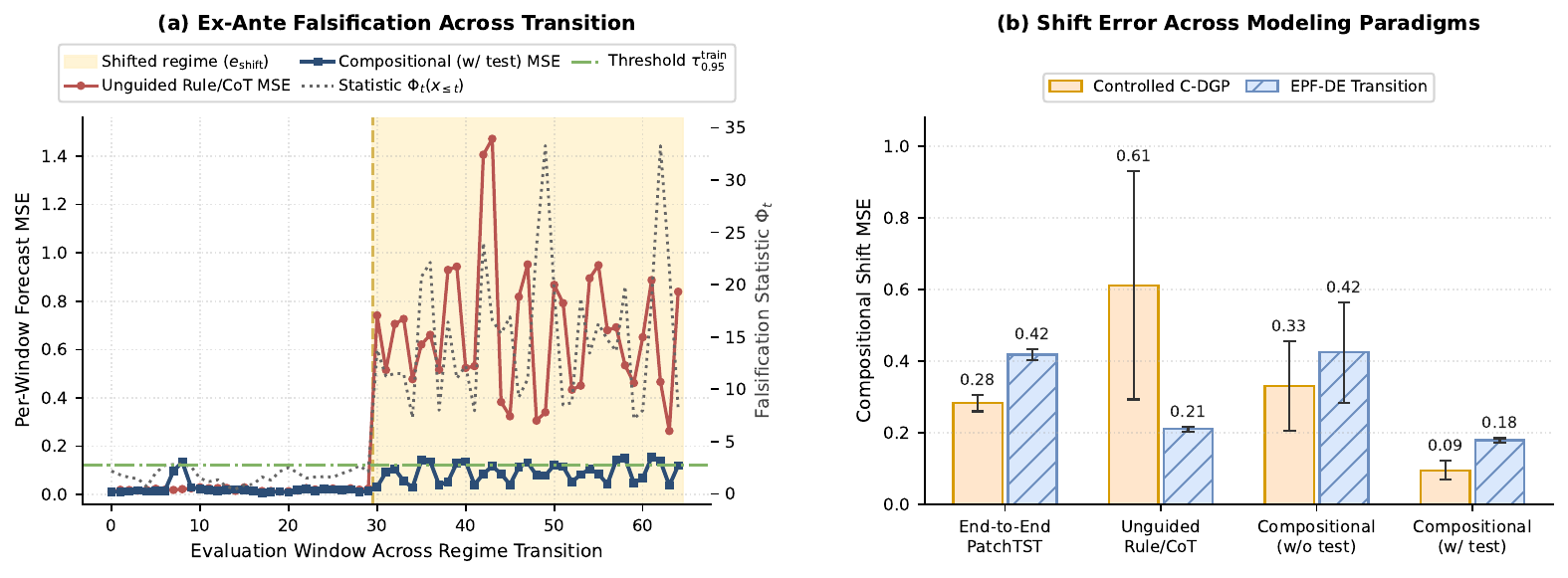}
  % \vspace{-0.6ex}
  \caption{\textbf{Ex-ante falsification testing versus unguided rule search and untested decomposition.} \textbf{(a)} Per-window forecast MSE and training-calibrated Mahalanobis statistic $\Phi_t(x_{\leq t})$ across a regime transition. When primitives decouple in $e_{\mathrm{shift}}$, $\Phi_t$ exceeds $\tau_{0.95}^{\mathrm{train}}$ and suppresses the invalidated residual pathway, preventing the error spikes seen in the backtest-selected rule chain. \textbf{(b)} Compositional-shift MSE (mean $\pm$ SD) across four modeling paradigms on both benchmarks.}
  \label{fig:probe2}
\end{figure*}

\subsection{Ex-Ante Falsification Testing Versus Unguided Rule Search}
\label{sec:emp_probe2}

Diagnostic Probe 2 evaluates whether symbolic rule search or primitive decomposition suffices without ex-ante falsification testing. Across five random seeds on both benchmarks (Figure~\ref{fig:probe2} and Table~\ref{tab:falsification}), we compare four methods under identical splits: (1) \textbf{End-to-End PatchTST} with covariates and RevIN, (2) \textbf{Unguided Rule/CoT}, which evaluates $M=120$ multi-step composite rules over lag differences, polynomial covariate interactions, and clock-hour terms to select the validation winner, (3) \textbf{Compositional (w/o test)}, which uses the same primitive decomposition and coupling operator as our full model while leaving its learned residual pathway active in all regimes, and (4) \textbf{Compositional (w/ test)}, which attenuates the residual pathway via a four-dimensional Mahalanobis test $\Phi_t(x_{\leq t}) \leq \tau_{0.95}^{\mathrm{train}}$ calibrated strictly on training inputs.

\begin{table*}[t]
  \centering
  \caption{In-distribution (\texttt{ID}) and compositional-shift (\texttt{Shift}) MSE (mean $\pm$ SD across 5 seeds), with test rejection rates (\%, \texttt{ID}/\texttt{Shift}).}
  \label{tab:falsification}
  % \vspace{0.3ex}
  % \scriptsize
  % \setlength{\tabcolsep}{2.2pt}
  \renewcommand{\arraystretch}{1}
  \begin{tabular}{lccc}
    \toprule
    \textbf{Method} & \textbf{ID MSE} $\downarrow$ & \textbf{Shift MSE} $\downarrow$ & \textbf{Reject \%} \\
    \midrule
    \multicolumn{4}{l}{\textit{Controlled Multi-Primitive Benchmark}} \\
    End-to-End PatchTST & $0.0439 \pm 0.0010$ & $0.2825 \pm 0.0232$ & $0.0 / 0.0$ \\
    Unguided Rule/CoT & $0.0198 \pm 0.0005$ & $0.6112 \pm 0.3185$ & $0.0 / 0.0$ \\
    Comp.\ (w/o test) & $\mathbf{0.0181 \pm 0.0008}$ & $0.3303 \pm 0.1260$ & $0.0 / 0.0$ \\
    Comp.\ (w/ test) & $0.0215 \pm 0.0006$ & $\mathbf{0.0948 \pm 0.0267}$ & $\mathbf{3.8 / 100}$ \\
    \midrule
    \multicolumn{4}{l}{\textit{German Electricity Transition}} \\
    End-to-End PatchTST & $0.2976 \pm 0.0126$ & $0.4174 \pm 0.0162$ & $0.0 / 0.0$ \\
    Unguided Rule/CoT & $\mathbf{0.1155 \pm 0.0007}$ & $0.2103 \pm 0.0067$ & $0.0 / 0.0$ \\
    Comp.\ (w/o test) & $0.1416 \pm 0.0080$ & $0.4239 \pm 0.1406$ & $0.0 / 0.0$ \\
    Comp.\ (w/ test) & $0.1410 \pm 0.0081$ & $\mathbf{0.1784 \pm 0.0058}$ & $\mathbf{0.5 / 98.4}$ \\
    \bottomrule
  \end{tabular}
\end{table*}

Table~\ref{tab:falsification} shows two clear patterns: (I) Unguided Rule/CoT achieves very low in-distribution MSE ($0.0198$ on \texttt{Controlled} and $0.1155$ on \texttt{EPF-DE}), yet its selected interaction terms fail after the shift, reaching $0.6112 \pm 0.3185$ on \texttt{Controlled} and $0.2103 \pm 0.0067$ on \texttt{EPF-DE}. Paired $t$-tests across seeds show lower shift error for Compositional (w/ test) relative to Unguided Rule/CoT on both \texttt{Controlled} ($0.0948 \pm 0.0267$, $t = -3.493, p = 0.0251$) and \texttt{EPF-DE} ($0.1784 \pm 0.0058$, $t = -6.386, p = 0.0031$). (II) Comparing Compositional (w/o test) with Compositional (w/ test) shows that primitive decomposition without falsification testing remains vulnerable to shift: when the local residual pathway trained on historical serial correlations is applied to novel primitive pairings, shift MSE increases from $0.0948$ to $0.3303$ on \texttt{Controlled} and from $0.1784$ to $0.4239$ on \texttt{EPF-DE}.

Figure~\ref{fig:probe2}a illustrates how the falsification test operates across the regime boundary using input statistics alone. During the stationary segment, $\Phi_t(x_{\leq t})$ remains below $\tau_{0.95}^{\mathrm{train}}$ ($3.8\%$ rejection rate on \texttt{Controlled} and $0.5\%$ on \texttt{EPF-DE}), keeping local residual refinements active. At the start of $e_{\mathrm{shift}}$, $\Phi_t$ rises above $\tau_{0.95}^{\mathrm{train}}$ ($100.0\%$ and $98.4\%$ rejection rates), shrinking the residual pathway and reverting to the invariant structural operator without requiring post-shift target labels.

\section{Formulating Falsifiable Compositional Time Series Models}
\label{sec:framework}
% =============================================================================

Using historical backtesting as a statistical test requires specifying structural assumptions before holdout evaluation and wiring neural sequence encoders to respect those assumptions at inference time.

\begin{table*}[t]
  \centering
  \caption{\textbf{Compositional Forecasting Hypothesis Card.} Before holdout evaluation, a falsifiable compositional forecast specifies five structural elements, illustrated for day-ahead electricity pricing (\texttt{EPF-DE}) and intraday financial order-flow returns.}
  \label{tab:hyp_card}
  \vspace{0.4ex}
  \footnotesize
  \renewcommand{\arraystretch}{1.16}
  \begin{tabularx}{\textwidth}{
    >{\bfseries\raggedright\arraybackslash}p{0.19\textwidth}
    >{\raggedright\arraybackslash}p{0.23\textwidth}
    >{\raggedright\arraybackslash}X
    >{\raggedright\arraybackslash}X
  }
    \toprule
    \textbf{Pre-Committed Element} &
    \textbf{Methodological Role} &
    \textbf{Example A: Day-Ahead Grid Pricing (\texttt{EPF-DE})} &
    \textbf{Example B: Intraday Order-Flow Returns} \\
    \midrule
    1. Grounded Primitives $\{m_k\}_{k=1}^K$ &
    Isolate reusable temporal drivers from entangled target lags. &
    $m_1$: 72h de-effected baseload level, $m_2$: grid load forecast, $m_3$: combined wind and solar supply. &
    $m_1$: dealer inventory imbalance, $m_2$: macro-announcement volatility state, $m_3$: cross-venue spread. \\
    \midrule
    2. Persistence Mechanism &
    Specify the physical or institutional mechanism sustaining the law. &
    Power-balance clearing along thermal merit-order dispatch curves \citep{weron2014electricity}. &
    Dealer inventory risk control requiring intraday liquidity compensation \citep{kyle1985continuous,cont2014price}. \\
    \midrule
    3. Directional Coupling $\mathcal{C}_{e_t}(\cdot)$ &
    Constrain signs, symmetries, or conservation relations across primitives. &
    Monotone positive response of $\Delta_{24} y_t$ to net residual load $\Delta_{24}(m_2 - m_3)$, steepening near reserve limits. &
    Negative return drift proportional to dealer inventory $m_1$, scaled inversely by depth $(1 + m_2)^{-1}$. \\
    \midrule
    4. Regime Boundary Conditions $e_t$ &
    Define observable states where $\mathcal{C}_{e_t}$ switches functional form or clips. &
    Curtailment and negative-price regime when renewable supply exceeds must-run thermal capacity. &
    Exchange price-limit halts, closing-auction cutoffs, and overnight margin liquidation windows. \\
    \midrule
    5. Ex-Ante Falsification Test $\Phi_t \leq \tau_\alpha^{\mathrm{train}}$ &
    Specify a training-calibrated support test that shrinks invalid residual fits. &
    4D Mahalanobis test on $m_2$, $m_3$, net load $m_2 - m_3$, and $\Delta_{24}(m_2 - m_3)$ attenuating diurnal lag residuals. &
    Rolling likelihood-ratio test on inventory-return sign consistency and spread jumps attenuating alpha residuals. \\
    \bottomrule
  \end{tabularx}
\end{table*}

\subsection{Pre-Committed Hypothesis Specification}
\label{sec:fw_card}

In quantitative finance, causal inference, and empirical econometrics, disciplined research protocols separate hypothesis formulation from holdout execution by fixing the mechanism, operating regime, and invalidation criteria of a model before inspecting out-of-sample performance \citep{popper1959logic,lopezdeprado2018advances,karlsson2025falsification}. Adapting this practice to machine learning yields a five-part specification (Table~\ref{tab:hyp_card}): (1) grounded temporal primitives $\{m_k\}_{k=1}^K$, (2) the physical, institutional, or behavioral persistence mechanism supporting them, (3) directional or conservation constraints on the coupling operator $\mathcal{C}_{e_t}$, (4) observable boundary conditions under which $\mathcal{C}_{e_t}$ clips or changes form, and (5) a training-calibrated ex-ante falsification statistic $\Phi_t(x_{\leq t}) \leq \tau_\alpha^{\mathrm{train}}$ paired with a fallback rule.

Table~\ref{tab:hyp_card} illustrates this specification across two domains with distinct noise characteristics. In day-ahead electricity forecasting (Instantiation A, evaluated in Section~\ref{sec:empirical}), separating slow baseload levels from net residual load $\mathrm{Load}_t - \mathrm{PVWind}_t$ encodes the physical balance condition that renewable output displaces thermal generation along the dispatch stack across both 5\% and 40\% renewable penetration. In intraday financial return prediction (Instantiation B), raw return autocorrelation is near zero and transient factor correlations decay quickly under arbitrage. Modelling dealer inventory imbalance ($m_1$) moderated by market depth ($m_2$) specifies both observable states where liquidity mean reversion suspends, such as price-limit halts or closing auctions, and an ex-ante test ($\Phi_t > \tau_\alpha^{\mathrm{train}}$) that zeros out residual predictions when order flow shifts toward directional informed trading \citep{kyle1985continuous,cont2014price}.

Elements 4 and 5 address complementary sources of non-stationarity. Element 4 (Regime Boundary Conditions) handles known, directly observable nonlinearities, such as capacity limits, weekend settlement rules, or statutory price caps, through deterministic changes in $\mathcal{C}_{e_t}$. Element 5 (Ex-Ante Falsification Test) handles unannounced shifts in the joint distribution of primitives before any discrete regime label changes. Testing whether the current input vector remains inside the training support of the residual pathway prevents local historical adjustments from extrapolating outside their calibration domain while retaining the primary coupling in Element 3.

\subsection{Coupling Pretrained Backbones with Ex-Ante Falsification Tests}
\label{sec:fw_arch}

In this architecture, deep sequence models and pretrained TSFMs parameterise the primitive encoders $\hat{m}_k = g_{\theta_k}(x_{\leq t}^{(k)})$ that extract structured latent signals from noisy sensor streams, clinical vitals, or grid series \citep{nie2023patchtst,ansari2024chronos,gu2024mamba,defelice2026interpretability}. Those primitive estimates pass into a constrained composition operator $\mathcal{C}_{e_t}(\hat{m}_{1:K})$, implemented via monotonic layers, conservation differences, or factorised energy terms \citep{du2024compositional}, paired with an attenuated residual pathway:
\begin{equation}
  \hat{y}_{t+1:t+H} = \mathcal{C}_{e_t}\!\bigl(\hat{m}_{1:K}\bigr) + \gamma\!\bigl(\Phi_t, \tau_\alpha^{\mathrm{train}}\bigr)\, r_\psi(x_{\leq t}),
  \label{eq:comp_arch}
\end{equation}
where $r_\psi(x_{\leq t})$ fits in-regime serial residuals and the shrinkage weight $\gamma(\Phi_t, \tau_\alpha^{\mathrm{train}}) \in [0, 1]$ decays toward zero when the test rejects ($\Phi_t(x_{\leq t}) > \tau_\alpha^{\mathrm{train}}$). Within stationary periods ($\gamma \approx 1$), Equation~\ref{eq:comp_arch} retains the expressive fit of end-to-end neural forecasters (Table~\ref{tab:falsification}). When primitives recombine outside training support ($\gamma \to 0$), the forecast reverts to $\mathcal{C}_{e_t}(\hat{m}_{1:K})$.

One possible role for LLMs is to help formulate candidate compositional hypotheses from domain knowledge instead of selecting forecasting rules solely by validation accuracy\citep{jin2024position,merrill2024language,williams2025context}. When LLMs map tokenised numbers directly to point forecasts or generate large pools of candidate chain-of-thought rules ranked on validation splits, evaluation remains vulnerable to the identifiability and multiple-testing limits in Section~\ref{sec:theory}. Used prior to backtesting, LLMs can translate domain documentation, market regulations, or clinical protocols into explicit candidate primitives, monotonicity constraints, and threshold specifications for Elements 1--5 of Table~\ref{tab:hyp_card}, while pretrained time series backbones estimate $\hat{m}_k$ and a single chronological holdout tests parameter fit.

\section{Conclusion}
\label{sec:conclusion}
% =============================================================================

Historical backtesting remains central to empirical time series modelling when used to evaluate explicit structural assumptions, yet treating chronological holdout error as an open-ended search and model-selection objective leaves collinear temporal primitives unidentifiable. Across both analytical derivations and diagnostic experiments on synthetic multi-primitive processes and German electricity market transitions, ranking unconstrained architectures, hyperparameter grids, or language-generated rule chains on finite historical splits systematically favours predictors tied to transient co-movement. Decomposing a system into grounded primitives is likewise insufficient on its own if unconstrained residual pathways continue extrapolating historical serial correlations after primitive relationships shift. 
Our analysis suggests that improving forecasting under structural shifts requires not only searching for better historical fits. A potential enhancement is to specify which temporal relationships should remain stable, test when learned residual corrections become unreliable, and restrict those corrections when their supporting assumptions no longer hold.
% Formulating time series forecasting and reasoning as falsifiable compositional hypotheses pairs neural sequence encoders and pretrained backbones with constrained coupling operators, observable regime boundaries, and training-calibrated ex-ante falsification tests that shrink invalid residual adjustments at inference time. Separating representation learning from ex-ante hypothesis specification aligns empirical evaluation with out-of-regime deployment across non-stationary environments.


\bibliography{references}
\bibliographystyle{icml2027}

%%%%%%%%%%%%%%%%%%%%%%%%%%%%%%%%%%%%%%%%%%%%%%%%%%%%%%%%%%%%%%%%%%%%%%%%%%%%%%%
%%%%%%%%%%%%%%%%%%%%%%%%%%%%%%%%%%%%%%%%%%%%%%%%%%%%%%%%%%%%%%%%%%%%%%%%%%%%%%%
% APPENDIX
%%%%%%%%%%%%%%%%%%%%%%%%%%%%%%%%%%%%%%%%%%%%%%%%%%%%%%%%%%%%%%%%%%%%%%%%%%%%%%%
%%%%%%%%%%%%%%%%%%%%%%%%%%%%%%%%%%%%%%%%%%%%%%%%%%%%%%%%%%%%%%%%%%%%%%%%%%%%%%%

%%%%%%%%%%%%%%%%%%%%%%%%%%%%%%%%%%%%%%%%%%%%%%%%%%%%%%%%%%%%%%%%%%%%%%%%%%%%%%%
%%%%%%%%%%%%%%%%%%%%%%%%%%%%%%%%%%%%%%%%%%%%%%%%%%%%%%%%%%%%%%%%%%%%%%%%%%%%%%%

\end{document}